\documentclass[11pt,a4paper]{article}
\usepackage{times}
\usepackage{latexsym}
\usepackage{amsmath,amsfonts}
\usepackage{tabularx}
\usepackage{booktabs} 
\usepackage{graphicx} 
\usepackage{emnlp2021}
\usepackage{multirow}

\usepackage{microtype}

\DeclareMathOperator*{\argmax}{arg\,max}

\newcommand{\vecX}{\mathbf{x}}
\newcommand{\vecY}{\mathbf{y}}
\newcommand{\vecZ}{\mathbf{z}}

\newcommand{\vecP}{\mathbf{p}}
\newcommand{\vecT}{\boldsymbol{\theta}}
\newcommand{\matP}{\mathbf{P}}

\newcommand{\changecolor}{FEA500}

\title{A Differentiable Language Model Adversarial Attack on Text Classifiers}

\author{Ivan Fursov$^1$, Alexey Zaytsev$^1$, Pavel Burnyshev$^{1,2}$, Ekaterina Dmitrieva$^3$, Nikita Klyuchnikov$^1$, \\ \bf Andrey Kravchenko$^4$, Ekaterina Artemova$^{2,3}$ and Evgeny Burnaev$^1$ \\

\href{mailto:A.Zaytsev@skoltech.ru}{A.Zaytsev@skoltech.ru} \\

$^1$ Skolkovo Institute of Science and Technology \\
$^2$ Huawei  Noah’s Ark lab \\
$^3$ HSE University \\
$^4$ Department of Computer Science, Oxford University 
}

\begin{document}
\maketitle
\begin{abstract}

% this abstract is already on arXiv, EA suggests rewriting 

% Steady progress and recent breakthroughs in natural language processing are mainly attributed to the development of contextualized encoders, which utilize efficient pre-training and powerful Transformer-based architectures and comprise billions of parameters. However, the gap in the performance between BERT, the most common model, and resent improved models is diminishing. 
% This may mean that the pre-trained contextualized encoders are reaching their limits. Efficient data usage, which leads to the performance increase, becomes a top priority. 

Robustness of huge Transformer-based models for natural language processing is an important issue due to their capabilities and wide adoption. 
One way to understand and improve robustness of these models is an exploration of an adversarial attack scenario: check if a small perturbation of an input can fool a model. 

Due to the discrete nature of textual data,  gradient-based adversarial methods, widely used in computer vision, are not applicable per~se. The standard strategy to overcome this issue is to develop token-level transformations, which do not take the whole sentence into account. 

In this paper, we propose a new black-box sentence-level attack. Our method fine-tunes a pre-trained language model to generate adversarial examples. A proposed differentiable loss function depends on a substitute classifier score and an approximate edit distance computed via a deep learning model. 

We show that the proposed attack outperforms competitors on a diverse set of NLP problems for both computed metrics and human evaluation. Moreover, due to the usage of the fine-tuned language model, the generated adversarial examples are hard to detect, thus current models are not robust. Hence, it is difficult to defend from the proposed attack, which is not the case for other attacks.

% An adversarial attack paradigm explores various scenarios for the vulnerability of deep learning models: minor changes of the input can force a model failure.  Over the last few years adversarial attack paradigm and adversarial training garnered enthusiasm from the NLP community. Transformations, ranging from morphology perturbations to swapping words with synonyms, can be applied to the input text to generate adversarial examples. A downstream model may benefit from training with the data augmented by adversarial examples. 

% Successful attacks on text classifiers are challenging since the model input are tokens from finite vocabulary, so a classifier score is non-differentiable with respect to inputs, and gradient-based attacks are not applicable. Common approaches deal with this problem by developing attacks at the token level.

% We instead fine-tune language model for adversarial attacks as a generator of adversarial examples. To optimize the model, we define a differentiable loss function that depends on a surrogate classifier score and on a deep learning model that evaluates approximate edit distance. So, we control both the adversability of a generated sequence and its similarity to the initial sequence.

\end{abstract}

\section{Introduction}
% Should we start with NLP tasks straightforwardly?

Adversarial attacks~\cite{yuan2019adversarial} in all application areas including computer vision~\cite{akhtar2018threat,khrulkov2018art}, natural language processing~\cite{zhang2019adversarial,wang2019survey,morris2020textattack}, and graphs~\cite{sun2018adversarial} seek to reveal non-robustness of deep learning models.
An adversarial attack on a text classification model perturbs the input sentence in such a way that the deep learning model is fooled, while the perturbations adhere to certain constraints, utilising morphology or grammar patterns or semantic similarity. The deep learning model then misclassifies the generated sentence, whilst for a human it is evident that the sentence's class remains the same~\cite{Kurakin2017}.
Unlike for images and pixels, for textual data it is not possible to estimate the derivatives of class probabilities with respect to input tokens due to the discrete nature of the language and its vocabulary. Although a sentence or a word representation can lie in a continuous space, the token itself can not be altered slightly to get to the neighbouring point. This turns partial derivatives useless.
Many approaches that accept the initial space of tokens as input attempt to modify these sequences using operations like addition, replacement, or switching of tokens~\cite{samanta2017towards, liang2017deep,ebrahimi2018hotflip}. Searching for the best modification can be stated as a discrete optimisation problem, which often appears to be computationally hard and is solved by random or greedy search heuristics in practice. \cite{ebrahimi2018hotflip}. Otherwise gradient optimisation techniques can be leveraged in the embedding space~\cite{sato2018interpretable,ren2020generating}.  This approach has a bottleneck:  all information contained in a sentence has to be squeezed into a single sentence embedding.

\begin{figure}
    \centering
    \includegraphics[width = 0.45\textwidth]{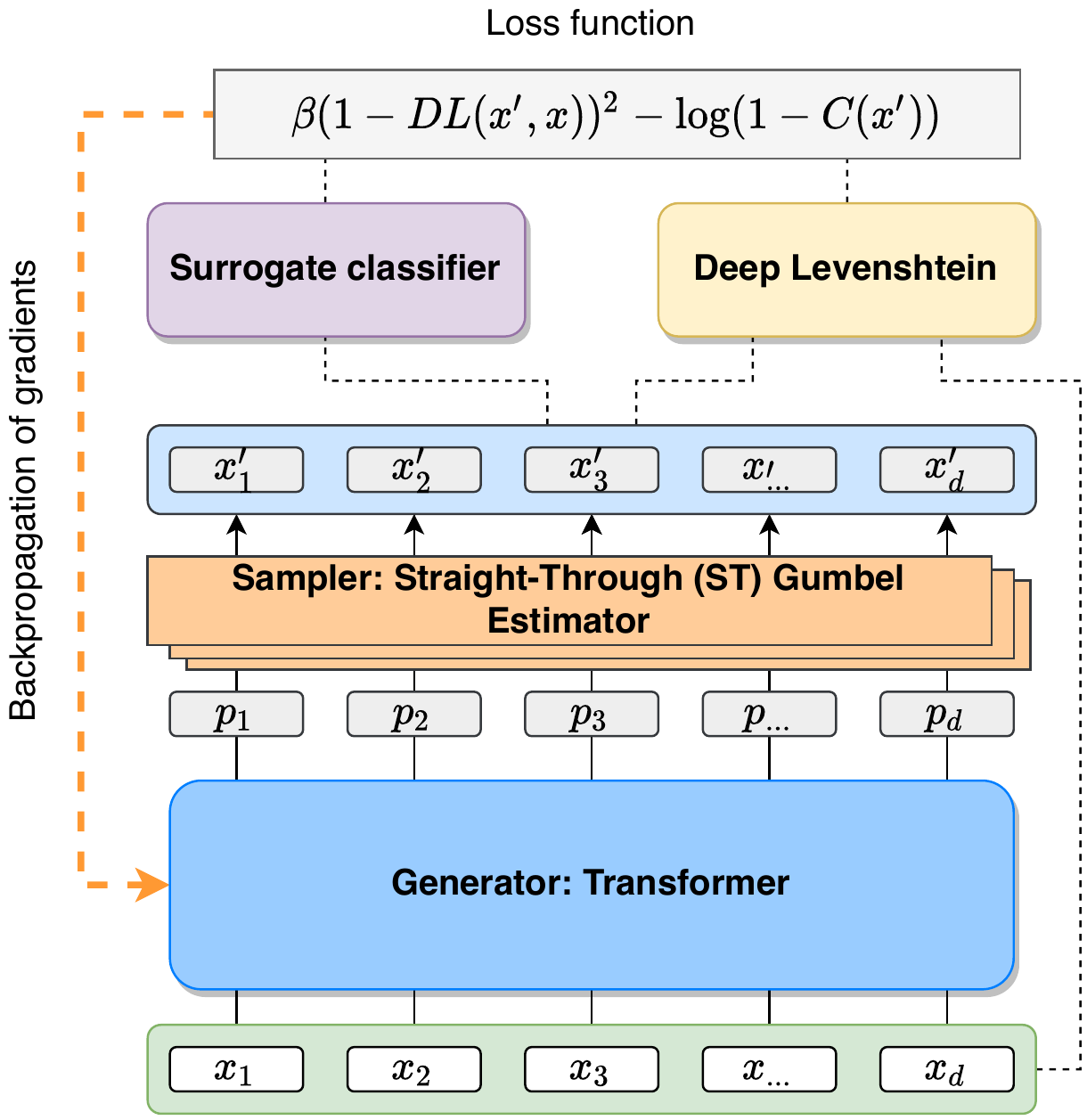}
  \caption{
  Training of the DILMA architecture consists of the following steps. \textbf{Step~1}: obtain logits $\matP$ from a pre-trained Language Model (LM) generator for input $\vecX$. \textbf{Step~2}: sample $\vecX'$ from the multinomial distribution~$\matP$ using the Gumbel-Softmax estimator. To improve generation quality we can sample many times. \textbf{Step~3}: obtain the substitute probability $C(\vecX')$ and approximate the edit distance $DL(\vecX', \vecX)$. \textbf{Step~4}: calculate the loss, do a backward pass. \textbf{Step~5}: update parameters of the LM using the computed gradients. }
    \label{fig:architecture}
\end{figure}

To alleviate these problems we propose a new differentiable adversarial attack model, which benefits from leveraging pre-trained language models, DILMA (DIfferentiable Language Model Attack).
The proposed model for an adversarial attack has two regimes. The first regime is a random sampling that produces adversarial examples by chance. The second regime is a targeted attack that modifies the language model by optimizing the loss with two terms related to misclassification by the target model and a discrepancy between an initial sequence and its adversarial counterpart.
Thus, we expect that the generated adversarial examples will fool a deep learning model, but will remain semantically close to the initial sequence.
% For the second regime we introduce a differentiable loss function as the weighted sum of the distance between the initial sequence and the generated one and the difference between the probability scores for these sequences.
We use a trained differentiable version of the Levenshtein distance~\cite{moon2018multimodal} and the Gumbel-Softmax heuristic to pass the derivatives through our sequence generation layers.
As our loss is differentiable, we can adopt any gradient-based adversarial attack.
The number of hyperparameters in our method is small, as we want our attack to be easily adopted to new datasets and problems.
The training and inference procedure is summarised in Fig.~\ref{fig:architecture}.
Examples of sentences generated via our attack DILMA are presented in Table~\ref{table:attack_samples}.

To summarise, the main contributions of this work are the following.
\begin{itemize}
    \item We propose a new black-box adversarial attack based on a masked language model (MLM) and a differentiable loss function to optimise during an attack. Our DILMA attack relies on fine-tuning parameters of an MLM by optimising the weighted sum of two differentiable terms: one is based on a surrogate distance between the source text  and its adversarial version, and another is a substitute classifier model score (Sec. \ref{sec:methods}). Hence, we adopt a generative model framework for the task of the generation of adversarial examples.
    \item We apply DILMA to various NLP sentence classification tasks and receive superior results in comparison to other methods (Sec. \ref{sec:experiments}).
    \item We show that a particular advantage of DILMA, when compared to other approaches, lies in the resistance to common defense strategies. The vast majority of  existing approaches fail to fool models after they defend themselves (Sec. \ref{sec:comparison}). 
    \item We provide a framework to extensively evaluate adversarial attacks for sentence classification and conduct a thorough evaluation of DILMA and related attacks. Adversarial training and adversarial detection must be an essential part of the adversarial attack evaluation on textual data in support to human evaluation  (Sec. \ref{sec:evaluation}).
\end{itemize}

\begin{table*}[h!]
\centering
\begin{tabular}{p{1.5cm}p{4.5cm}p{4.5cm}p{4.5cm}}
\toprule
Dataset & Original& PWWS & DILMA~(ours) \\
\midrule
%rotten_tomatoes
rotten tomatoes & flat, misguided comedy & 
\!\!
{\color[HTML]{\changecolor}unconditional}, {\color[HTML]{\changecolor} misadvise} comedy 
& 
\phantom{$a$}\!\!\!\!
{\color[HTML]{\changecolor} witty} and {\color[HTML]{\changecolor} misguide good} comedy
\\  
\cline{2-4} 
 & more likely to have you scratching your head than hiding  under your seat & 
more {\color[HTML]{\changecolor}potential} to have you {\color[HTML]{\changecolor}scraping} your {\color[HTML]{\changecolor}forefront} than hiding under your seat & 
more likely to have you scratching your head than {\color[HTML]{\changecolor}brood beneath} your seat \\   \cline{2-4} 
 & a rehash of every gangster movie from the past decade & a {\color[HTML]{\changecolor}retrograde} of every gangster movie from the past decade & 
a rehash of every {\color[HTML]{\changecolor}good film} {\color[HTML]{\changecolor}of} the past decade \\ \hline 

%sst2

sst2 &this is a story of two misfits who dont stand a chance alone but together they are magnificent & this is a {\color[HTML]{\changecolor}floor} of two misfits who dont {\color[HTML]{\changecolor}stall} a chance alone but {\color[HTML]{\changecolor}unitedly} they are magnificent & this is a story of two misfits who do not stand a chance alone but together they are {\color[HTML]{\changecolor}united} \\ \cline{2-4} 
& the draw for big bad love is a solid performance by arliss howard & the draw for {\color[HTML]{\changecolor}gravid} bad love is a {\color[HTML]{\changecolor}square} performance by arliss howard & the song for big bad love is a {\color[HTML]{\changecolor}1964} performance by arliss howard \\ \cline{2-4} 
& a gripping movie played with performances that are all understated and touching & a gripping movie played with performances that are all {\color[HTML]{\changecolor}downplay} and {\color[HTML]{\changecolor}advert} & a {\color[HTML]{\changecolor}horror} movie with {\color[HTML]{\changecolor}lyrics} that are all understated and {\color[HTML]{\changecolor}demanding} \\ \hline

%dstc

dstc & i wanna play the movie & i wanna {\color[HTML]{\changecolor}frolic} the movie & i wanna play the {\color[HTML]{\changecolor}games} \\ \cline{2-4} 
& could you help me call a cab to the airport? & could you help me {\color[HTML]{\changecolor}visit} a {\color[HTML]{\changecolor}cabriolet} to the airport?  & could you help me find a {\color[HTML]{\changecolor}taxi} to the airport? \\ \cline{2-4} 

& i would like to book 1 room at a nearby hotel. & i would like to book {\color[HTML]{\changecolor}single} room at a nearby hotel. & i would like to book 1 room at a {\color[HTML]{\changecolor}approved} hotel. \\

\bottomrule 
\end{tabular}
\caption{Attack samples for the PWWS attack (top performing according to our evaluation) and our attack DILMA. DILMA can provide more diverse adversarial sequences with meanings similar to that of the initial sentence.}
\label{table:attack_samples}
\end{table*}

\section{Related work} 

There exist adversarial attacks for different types of data:
the most mainstream being image data~\cite{szegedy2014intriguing,goodfellow2014explaining}, graph data~\cite{zugner2018adversarial}, and sequences~\cite{papernot2016crafting}. The latter work is one of the first publications on generation of adversarial attacks for discrete sequences, such as texts. It identifies two main challenges for the task: a discrete space of possible objects and a complex definition of a semantically coherent sequence.

There is a well-established categorisation of textual attacks according to the perturbation level. 
{\bf Character-level} attacks include replacing some of the characters with visually similar ones \cite{eger2019text}. 
HotFlip uses gradient-based decisions to swap, remove, or insert characters \cite{ebrahimi2018hotflip}. \cite{tan2020s} creates sub-word perturbations to craft non-standard English adversaries and mitigates the cultural biases of machine translation models.  
{\bf Token-level} attacks replace a token (mostly word) in a given sentence, based on a salience score and following a specific synonym replacement strategy.  \citet{meng2020geometry} choose the best synonym replacement from WordNet to generate a sentence, close to the classifier's decision boundary. \citet{li2020bert} and \citet{garg2020bae} utilise BERT's masked model to generate the substitution to the target word in multiple ways. The Metropolis-Hastings algorithm improves sampling from a constrained distribution, allowing to substitute a word in a sentence which belongs to a desired class \cite{zhang2019generating}. 
{\bf Sentence-level} attacks aim to generate a new adversarial instance from scratch with the help of paraphrasing models \cite{gan2019improving}, back translation \cite{zhang2019paws} or competitive dialogue agents  \cite{cheng2019evaluating}.

Another way to categorise textual adversarial attacks accounts for the amount of information received from the victim target model.  
{\bf White box} attacks have full access to the victim model and its gradients. Generating adversarial examples guided by the training loss is intractable due to the discrete nature of textual data. Instead greedy methods \cite{cheng2019robust} or the Gumbel trick \cite{yang2020greedy} are used to make computations tractable. 
{\bf Black box} attacks have access only to the model outputs. Victim's scores may help to estimate word salience \cite{jin2020bert}.  Black box attacks can emulate white box attacks. For example, DistFlip \cite{gil2019white} is a faster version of HotFlip with a similar accuracy.

{\bf Blind} attacks do not have access to the target model at all and can be seen as a form of text augmentation. Such attacks include stop word removal, dictionary-based word replacement, and rule-based insertion of negation \cite{niu2018adversarial}. 

Methods to defend against attacks exploit adversarial stability training \cite{liu2020joint} and robust word and character representations \cite{jones2020robust}. Back-off strategies to recognise intentionally replaced and corrupted words can detect perturbed inputs before actually passing them through a classifier \cite{pruthi2019combating}.  

Common tasks requiring defence from adversarial attacks by adversarial training, include
classification, sentence pair problems, such as natural language inference (NLI) \cite{jin2020bert}, and machine translation \cite{cheng2020seq2sick,huang2020reinforced}. 

% A noteworthy related research direction includes methods for model extraction \cite{keskar2020thieves}. A model, extracted from an industrial setting, such as machine translation API, may be prone to the same attacks, as the source model. To protect the model, \citet{wallace2020imitation} develop ensemble decoding algorithms that prevent from model extraction. 

From a technical point of view, two open source frameworks for textual adversarial attacks, adversarial training, and text augmentation, namely, \texttt{OpenAttack} \cite{zeng2020openattack} and \texttt{TextAttack} \cite{morris2020textattack},  facilitate the research in the area and help to conduct fast and correct comparison. 

From the current state of the art, we see a lack of effective ways to generate adversarial categorical sequences and defend from such attacks.
Existing approaches use previous generations of LMs based on recurrent architectures, stay at a token level, or use VAEs, despite known limitations of these models for modelling NLP data~\cite{yang2017improved,vaswani2017attention}. %Moreover, as most of the applications focus on NLP-related tasks, it makes sense to widen the scope of application domains for adversarial attacks on categorical sequences. 

\section{Methods} %Alexey, Ivan
\label{sec:methods}

\subsection{General description of the approach} 
We generate adversarial examples using two consecutive components: a masked language model (MLM) with parameters $\vecT$ that provides for an input sequence $\vecX$, conditional distribution $p_{\vecT}(\vecX'|\vecX)$, and a sampler from this distribution such that $\vecX' \sim p_{\vecT}(\vecX'|\vecX)$.
Thus, we can generate sequences $\vecX'$ by a consecutive application of the MLM and the sampler.

For this sequence to be adversarial, we optimise a proposed differentiable loss function that forces the MLM to generate semantically similar but adversarial examples by modifying the MLM parameters $\vecT$.
The loss function consists of two terms: the first term corresponds to a substitute classifier $C(\vecX')$ that outputs a probability of belonging to a target class, and the second term corresponds to a Deep Levenshtein distance $DL(\vecX, \vecX')$ that approximates the edit distance between sequences.

The general scheme of our approach is given in Fig.~\ref{fig:architecture}.
We start with the description of the LM and the sampler in Subsection~\ref{sec:seq2seqmodel}.
We continue with the description of the loss function in Subsection~\ref{sec:loss_function}.
The formal description of our algorithm is given in Subsection~\ref{sec:dilma_description}.
In later subsections~\ref{sec:classifiers} and~\ref{sec:deep_levenstein}, we provide more details on the use of the target and substitute classifiers and the Deep Levenshtein model correspondingly.
% Detailed descriptions of the architectures and training procedures is provided in supplementary materials~\ref{sec:classifiers_sup}.

% consider rewriting:
% Subsection~\ref{sec:seq2seqmodel} describes the MLM and the sampler. Subsection~\ref{sec:loss_function} presents the loss function. Subsection~\ref{sec:dilma_description} provides with the formal description of algorithm, while  Subsections~\ref{sec:classifiers} and~\ref{sec:deep_levenstein} show the use of the target and surrogate classifiers and the Deep Levenstein model in more details. 

\subsection{Masked Language model} 
\label{sec:seq2seqmodel}

%The masked language model (MLM) lies at the heart of our approach.
The MLM is a model that takes a sequence of tokens (e.g. words) as input $\vecX = \{x_1, \ldots, x_t\}$ and outputs logits for tokens $\matP = \{\vecP_1, \ldots, \vecP_{t}\} \in \mathbb{R}^{d \cdot t}$ for each index $1, \ldots, t$, where $d$ is a size of the dictionary of tokens. 
In this work we use a transformer architecture as the LM~\cite{vaswani2017attention}.
We pre-train a transformer encoder \cite{vaswani2017attention} in a BERT \cite{devlin2018bert} manner from scratch using the available data. 
% We use all available data to train such this kind of model.

The sampler is defined as follows: the MLM learns a probability distribution over sequences, so we can sample a new sequence $\vecX' = \{x'_1, \ldots, x'_t\}$ based on the vectors of token logits $\matP$. In this work we use Straight-Through Gumbel Estimator $ST(\matP): \matP \rightarrow \vecX'$ for sampling~\cite{jang2017categorical}.
To get the actual probabilities $q_{ij}$ from logits $p_{ij}$, we use a softmax with temperature $\tau$ \cite{hinton2015distilling}, where $\tau > 1$ produces a softer probability distribution over classes. As $\tau \rightarrow \infty$, the original distribution approaches a uniform distribution. If $\tau\to0$, then sampling from it reduces to $\argmax$ sampling.

% setting $x'_i = \argmax_j \vecP_i = \argmax \{p_{i1}, \ldots, p_{id}\}^{\top}$.
%     \begin{equation}
%     \label{eqeq}
%         q_{ij} = \frac{\exp \left(p_{ij} / \tau \right)}{\sum_{k = 1}^d \exp \left(p_{ik} / \tau \right)}.
%     \end{equation}
% The value $\tau > 1$ produces a softer probability distribution over classes. As $\tau \rightarrow \infty$, the original distribution approaches a uniform distribution. If $\tau\to0$, then sampling from \eqref{eqeq} reduces to setting $x'_i = \argmax_j \vecP_i = \argmax \{p_{i1}, \ldots, p_{id}\}^{\top}$.

Our first method \textbf{SamplingFool} samples sequences from the categorical distribution with $q_{ij}$ probabilities. 
%The sampled examples turn out to be good-looking and can easily fool a classifier, as we discuss in Section \ref{sec:experiments}. 
The method is similar to the random search algorithm and serves as a baseline.
%for the generation of adversarial examples for discrete sequences.

\subsection{Loss function}
\label{sec:loss_function}

The Straight-Through Gumbel sampling allows a propagation of the derivatives through the softmax layer.
Hence, we optimise parameters $\vecT$ of our MLM to improve the quality of generated adversarial examples.

Let $C^t_y(\vecX)$ be the probability of the target classifier to predict the considered class $y$/. A loss function takes two terms into account: the first term estimates the probability score drop $(1 - C^t_y(\vecX'))$ of the target classifier, the second term represents the edit distance $DL(\vecX, \vecX')$ between the initial sequence $\vecX$ and the generated sequence $\vecX'$. 
We should maximise the probability drop and minimise the edit distance, so it is as close to $1$ as possible.

In the black-box scenario we do not have access to the true classifier score, so we use a substitute classifier score $C_y(\vecX') \approx C^t_y(\vecX')$.
% More details on classifiers are given in Subsection~\ref{sec:classifiers}.
As a differentiable alternative to edit distance, we use the Deep Levenshtein model proposed in \cite{moon2018multimodal} --- a deep learning model that approximates the edit distance: $DL(\vecX, \vecX') \approx D(\vecX, \vecX')$.
% More details on the Deep Levenstein model are given in Subsection~\ref{sec:deep_levenstein}.

This way, the loss function becomes:
\begin{equation}\label{eq:dilma_loss}
    L(\vecX', \vecX, y) = \beta (1 - DL(\vecX', \vecX))^2 - \log (1 - C_y(\vecX')), 
\end{equation}
where $C_y(\vecX')$ is the probability of the true class $y$ for sequence $\vecX'$ and $\beta$ is a weighting coefficient. Thus, we penalise cases when a modification in more than one token is needed in order to get $\vecX'$ from $\vecX$. Since we focus on non-target attacks, the $C_y(\vecX')$ component is included in the loss. The smaller the probability of an attacked class, the smaller the loss.

We estimate derivatives of $L(\vecX', \vecX, y)$ in a way similar to~\cite{jang2016categorical}.
Using these derivatives, a backward pass updates the weights $\vecT$ of the MLM. We find that updating the whole set of parameters $\vecT$ is not the best strategy, and a better alternative is to update only the last linear layer and the last layer of the generator.

We consider two DILMA options: initial \textbf{DILMA} that minimises only the classifier score (the second term) during the attack and \textbf{DILMA with DL} that takes into account the approximate Deep Levenshtein edit distance.

\subsection{DILMA algorithm}
\label{sec:dilma_description}

Now we are ready to group the introduced components into a formal algorithm with the architecture depicted in Fig.~\ref{fig:architecture}.

% The proposed approach for the generation of adversarial sequences has the following inputs:
% a language model with parameters $\vecT = \vecT_0$, learning rate $\alpha$, temperature for sampling $\tau$, coefficient $\beta$ (if $\beta = 0$, we only decrease a classifier score and don't pay attention to the edit distance part of the loss function) for the normalised classifier score in~\eqref{eq:dilma_loss}.
% For these inputs the algorithm performs the following steps at iterations $i = 1, 2, \ldots, k$ for a given sequence $\vecX$.

Given a sequence $\vecX$, the \textbf{DILMA} attack performs the following steps at iterations $i = 1, 2, \ldots, k$ starting from the parameters of a pre-trained MLM $\vecT_0 = \vecT$ and running for $k$ iterations.
\begin{itemize}
    \item[\textbf{Step 1.}] Pass the sequence $\vecX$ through the pre-trained MLM. Obtain logits $\matP = LM_{\vecT_{i - 1}}(\vecX)$.
    \item[\textbf{Step 2.}] Sample an adversarial sequence $\vecX'$ from the logits using the Gumbel-Softmax Estimator.
    \item[\textbf{Step 3.}] Calculate the probability $C_y(\vecX')$ and the Deep Levenshtein distance $DL(\vecX', \vecX)$. Calculate the loss value $L(\vecX', \vecX, y)$~\eqref{eq:dilma_loss}.
    \item[\textbf{Step 4.}] Do a backward pass to update the LM's weights $\vecT_{i - 1}$ using gradient descent and get the new weights $\vecT_{i}$.
    \item[\textbf{Step 5.}] Obtain an adversarial sequence $\vecX'_i$ by sampling based on~the softmax with a selected temperature.
        %with $\tau > 0$ and $\matP$.
\end{itemize}

The algorithm chooses which tokens should be replaced.  The classification component changes the sequence in a direction where the probability score $C_y(\vecX')$ is low, and the Deep Levenshtein distance keeps the generated sequence close to the original one.
%The update procedure for each $\vecX$ starts from the pre-trained initial LM parameters $\vecT_0$. 

%  \textbf{DILMA w/ sampling} samples  $m>1$  adversarial examples for each iteration with almost no additional computational cost. As we will see in Section \ref{sec:experiments}, this approach works better than the original \textbf{DILMA} approach across all datasets.

\begin{table*}[]
    \centering
    \begin{tabular}{cccccccc}
    \toprule
& Classes & Avg.  & Max  & Train  & Test & Targeted RoBERTa & Substitute LSTM \\
& & length & length & size & size & accuracy & accuracy \\
\midrule
AG & 4 & 6.61 & 19 & 115000 & 5000 & 0.87 & 0.86 \\
DSTC & 6 & 8.82 & 33 & 5452 & 500 & 0.85 & 0.85 \\
SST-2 & 2 & 8.62 & 48 & 62349 & 5000 & 0.82 & 0.80 \\
RT & 2 & 18.44 & 51 & 8530 & 1066 & 0.76 & 0.70 \\
\bottomrule
    \end{tabular}
    \caption{Description of the datasets}
    \label{table:datasets_comparison}
\end{table*}

% After all iterations $i = 1, 2, \ldots, k$ of the algorithm we obtain a set of adversarial sequences $\{\vecX_i'\}_{i = 1}^k$ in case of the original \textbf{DILMA}; for \textbf{DILMA w/ sampling} we get $m$ adversarial sequences on each iteration of the algorithm. 
We obtain a set of $m$ sampled adversarial sequences on each iteration of the algorithm. 
The last sequence in this set is not always the best one. Therefore among generated sequences that are adversarial w.r.t. the substitute classifier $C_y(\vecX')$ we choose $\vecX'_{\mathrm{opt}}$ with the lowest  Word Error Rate (WER), estimated with respect to the initial sequence $\vecX$.
If all examples don't change the model prediction w.r.t. $C_y(\vecX')$, then we select $\vecX'_{\mathrm{opt}}$ with the smallest target class score, estimated by the substitute classifier. 

We choose the hyperparameter set achieving the best NAD score with the Optuna framework \cite{optuna_2019}.

% As we use only the substitute classifier, \textbf{DILMA} is a blackbox-type adversarial attack.

% \begin{algorithm}
%  \KwIn{Pre-trained $LM$ with parameters $\theta_{LM}$, substitute classifier $C(x)$, Deep Levenshtein model $DL(x', x)$, number of steps $N$}
%  \KwData{Original sequence $x$ and true label $y \in \{1, \dots, k\}$}
%  \KwResult{Candidate adversarial sequences $\{x'_i\}_{i = 1}^N$}
%  \For{$i \gets 1$ \textbf{to} $N$} {
%   $p_i := LM(x)$\;
%   $x'_i := Gumbel(p_i)$ is an adversarial sequence\;
%   $C_y(x'_i)$ is a probability of class $y$\;
%   $DL(x'_i, x)$ is an approximated WER\;
%   $L_i = (1 - DL(x'_i, x))^2 - \alpha \log (1 - C_y(x'_i))$ \;
%   $\theta_{LM} := \theta_{LM} - \beta \delta L_i$ is update of LM parameters\;
%   $x'_i = \arg \max LM(x)$
%  }
%  \caption{DILMA Attack Algorithm}
%  \label{alg:attack_pseudocode}
% \end{algorithm}

\subsection{Classification model}
\label{sec:classifiers}

In all experiments we use two classifiers: a target classifier $C^t(\vecX)$ that we attack and a substitute classifier $C(\vecX)$ that provides differentiable substitute classifier scores. 

The target classifier is RoBerta \cite{liu2019roberta}. The substitute classifier is the LSTM model. The hidden size is $150$, the dropout rate is $0.3$.

% The target classifier is a combination of a bi-directional Gated Recurrent Unit (GRU)~\cite{chung2014empirical} RNN and an embeddings layer for input tokens. 
% The hidden size for GRU is $128$, the dropout rate is $0.1$ and the embedding size is $100$.
% The substitute classifier is a Convolutional Neural Network (CNN) for sentence classification~\cite{kim2014convolutional} and an embeddings layer of size $100$ for tokens before it.

The substitute classifier has access only to $50\%$ of the data, whilst the target classifier uses the entire dataset. We split the dataset into two parts keeping the class balance similar for each part. 

\subsection{The Deep Levenshtein Model}
\label{sec:deep_levenstein}

The differentiable version of the edit distance allows gradient-based updates of parameters. 
Following the Deep Levenshtein approach \cite{moon2018multimodal}, we train a deep learning model $DL(\vecX, \vecX')$ to estimate the  Word Error Rate (WER) between two sequences $\vecX$ and $\vecX'$. 
We treat WER as the word-level Levenshtein distance.

% distance, since we work at the word level instead of the character level for NLP tasks, and for non-textual tasks there are no levels other than ``tokens''.

Following \cite{dai2020convolutional}, the Deep Levenshtein model receives two sequences $(\vecX, \vecY)$.  It encodes them into dense representations $\vecZ_\vecX = E(\vecX)$ and $\vecZ_\vecY = E(\vecY)$ of fixed length $l$ using the shared encoder. Then it concatenates the representations and the absolute difference between them in a vector $(\vecZ_\vecY, \vecZ_\vecY, |\vecZ_\vecY - \vecZ_\vecY|, \vecZ_ * \vecY)$ of length $4l$. 
At the end the model uses a fully-connected layer to predict WER.
% The architecture is similar to the one proposed in~\cite{dai2020convolutional}. 
To estimate the parameters of the encoder and the fully connected layer we use the $L_2$ loss between the true and the predicted WER values.
We form a training sample of size of about two million data points by sampling pairs of sequences and their modifications from the training data.

\begin{table*}[]
\centering
\begin{tabular}{l|c|c|c|c}
\toprule
\multicolumn{1}{l|}{} & \textbf{AG}   & \textbf{DSTC}   & \textbf{SST-2}     & \textbf{RT} \\ \midrule
DeepWordBug          & 0.017 / \textbf{0.012}          & 0.156 / 0.034          & 0.12 / 0.077           & \underline{0.071} / 0.054          \\
HotFlip              & 0.015 / 0.009          & 0.09 / 0.084           & 0.023 / 0.001          & 0.034 / 0.011          \\
PWWS                 &  0.018 / \underline{0.011}    & 0.142 / 0.025          & 0.156 / 0.068          & \textbf{0.102} / \textbf{0.072} \\
TextBugger           & 0.015 / 0.009          & 0.098 / 0.023          & 0.066 / 0.041          & 0.066 / \underline{0.059}          \\
\textit{SamplingFool (ours)}  & 0.015 / 0.004          & 0.218 / \underline{0.113}          & 0.137 / 0.126          & 0.04 / 0.022           \\
\textit{DILMA (ours)}         & \textbf{0.023} / 0.003 & \underline{0.237} / 0.111          & \underline{0.19} / \underline{0.154}           & 0.045 / 0.016          \\
\textit{DILMA with DL (ours)} & \underline{0.022} / 0.004    & \textbf{0.241} / \textbf{0.131} & \textbf{0.208} / \textbf{0.172} & 0.052 / 0.02           \\ \hline
\end{tabular}
\caption{NAD metric ($\uparrow$) before/after adversarial training on 5,000 examples. The best values are in \textbf{bold}, the second best values are \underline{underscored}. 
DILMA is resistant to adversarial training as well as Sampling Fool.}
\label{table:main_results}
\end{table*}

% ================================================================================
\section{Experiments} %Ivan, Pavel
\label{sec:experiments}

The datasets and the source code are published online\footnote{The code is available at \url{https://anonymous.4open.science/r/4cf31d59-9fe9-4854-ba53-7be1f9f6be7d/}}.

% An ablation study and an algorithm to select hyperparameters as well as additional experiments are provided in supplementary materials in Section~\ref{sec:additional_experiments}.

\subsection{Competing approaches}

We compared our approach to other popular approaches described below. 
%HotFlip, FGSM variant for adversarial sequences~\cite{goodfellow2014explaining, papernot2016crafting}, and DeepFool~\cite{moosavidezfooli2015deepfool}.
% We have also tried to implement~\cite{ren2020generating}, but haven't managed to find hyperparameters that provide performance similar to that reported by the authors. 

\textbf{HotFlip}~\cite{ebrahimi2018hotflip} selects the best token to change, given an approximation of partial derivatives for all tokens and all elements of the dictionary. To change multiple tokens HotFlip selects the sequence of changes via a beam search.

% \textbf{FGSM}~\cite{goodfellow2014explaining} chooses a random token in a sequence and uses the Fast Gradient Sign Method to perturb its embedding vector. Then the algorithm finds the closest vector in an embedding matrix and replaces the chosen token with the one that corresponds to the identified vector.

% \textbf{DeepFool}~\cite{moosavidezfooli2015deepfool} follows the same idea of a gradient-based replacement method in an embedded space.
% In addition, by assuming the local linearity of a classifier, DeepFool provides a heuristic to select the most promising modification. 

\textbf{Textbugger}~\cite{li2019textbugger} works at a symbol level and tries to replace symbols in words to generate new adversarial sequences using derivative values for possible replacements.

\textbf{DeepWordBug}~\cite{gao2018black} is a greedy replace-one-word heuristic scoring based on an estimate of importance of a word to a RNN model with each replacement being a character-swap attack. DeepWordBug is a black box attack.

\textbf{PWWS}~\cite{ren-etal-2019-generating} is a greedy synonym-swap method, which takes into account the saliencies of the words and the effectiveness of their replacement. Replacements of the words are made with the help of the WordNet dictionary. 

\subsection{Datasets}

We have conducted experiments on four open NLP datasets for different tasks such as text classification, intent prediction, and sentiment analysis. The characteristics of these datasets are presented in Table~\ref{table:datasets_comparison}.

% To normalise all NLP data, we lowercase the text and drop all non-alphabetic characters (e.g. numbers) and special symbols. 
\textbf{The AG News corpus} (AG) \cite{zhang2015characterlevel} consists of news articles on the web from the AG corpus. There are four classes: World, Sports, Business, and Sci/Tech. Both training and test sets are perfectly balanced.
The \textbf{Dialogue State Tracking Challenge} dataset (DSTC) is a special processed dataset related to dialogue system tasks. The standard DSTC8 dataset \cite{rastogi2020schema} was adopted to the intent prediction task by extracting most intent-interpreted sentences from dialogues. 
\textbf{The Stanford Sentiment Treebank} (SST-2) \cite{socher-etal-2013-recursive} contains phrases with fine-grained sentiment labels in the parse trees of $11,855$ sentences from movie reviews.
\textbf{The Rotten Tomatoes} dataset (RT) \cite{PangLee2005Seeing} is a film-review dataset of sentences with positive or negative sentiment labels.

\subsection{NAD metric}

To create an adversarial attack, changes must be applied to the initial sequence. A change can be done either by inserting, deleting, or replacing a token in some position in the original sequence. In the $WER$ calculation, any change to the sequence made by insertion, deletion, or replacement costs $1$. Therefore, we consider the adversarial sequence to be perfect if $WER = 1$, and the target classifier output has changed. For the classification task, Normalised Accuracy Drop (NAD) is calculated in the following way:
\[
    NAD(A) = \frac{1}{N} \sum_{i=1}^N \frac{\mathbf{1}\{ C^t(\vecX_i) \neq C^t(\vecX'_i)) \}}{WER(\vecX_i, \vecX'_i)},
\]
where $\vecX' = A(\vecX)$ is the output of an adversarial generation algorithm for the input sequence $\vecX$, $C^t(\vecX)$ is the label assigned by the target classification model, and $WER(\vecX', \vecX)$ is the Word Error Rate. The highest value of NAD is achieved when $WER(\vecX_i', \vecX_i) = 1$ and $C(\vecX_i) \neq C(\vecX'_i)$ for all $i$. Here we assume that adversaries produce distinct sequences and $WER(\vecX_i, \vecX'_i) \geq 1$.

\section{Evaluation and comparison}

\subsection{Adversarial attack evaluation}
\label{sec:comparison}

Results for the considered methods are presented in Table~\ref{table:main_results}.
We demonstrate not only the quality of attacks on the initial target classifier, but also after its re-training with additional adversarial samples added to the training set.
After re-training the initial target classifier, competitor attacks cannot provide reasonable results, whilst our methods perform significantly better before and after retraining for most of the datasets.
% In case of the \textbf{Tr.Age} dataset, SamplingFool works better than DILMA because of the overall low quality of the target classifier.
We provide additional attack quality metrics in supplementary materials.

\subsection{Additional metrics for evaluation}
\label{sec:comparison}

We provide two additional metrics to judge the quality of our and competitors' attacks:
\textbf{accuracy} of target models changing after attacks; \textbf{probability difference} is a difference between true model scores before and after an attack.
Higher values for \textbf{probability differences} mean that an attack is more successful in affecting the target model decision.

Discriminator training~\cite{xu2019adversarial} is another defence strategy we followed. 
We trained an additional discriminator on $10,000$ samples of the original sequences and adversarial examples. The discriminator detects whether an example is normal or adversarial.
The discriminator has an LSTM architecture with fine-tuning of GLoVE embeddings and was trained for $50$ epochs using the negative log-likelihood loss with early stopping based on validation scores.

We provide all these scores in Table~\ref{table:accuracy_table}.
As we can see our methods are top performer with respect to the obtained accuracy scores and probability differences.
Due to smaller and more careful changes, HotFlip is harder to detect by a discriminator.
However, ROC AUC scores are close to a ROC AUC score of a random classifier for two of four datasets for our attacks.

%Accuracy
\begin{table*}[ht]
\small
\centering
\begin{tabular}{lcccccccccccc}
\toprule
\multicolumn{1}{l}{} & \multicolumn{4}{c}{Accuracy score} & \multicolumn{4}{c}{PD metric}
& \multicolumn{4}{c}{ROC AUC for} \\
\multicolumn{1}{l}{} & \multicolumn{4}{c}{} & \multicolumn{4}{c}{} & \multicolumn{4}{c}{Adv.discriminator} \\
\cmidrule(r){2-5} \cmidrule(r){6-9} \cmidrule(r){10-13}
\multicolumn{1}{l}{} & \textbf{AG}  & \textbf{DSTC}   & \textbf{SST-2}     & \textbf{RT} & \textbf{AG}  & \textbf{DSTC}   & \textbf{SST-2}     & \textbf{RT} & \textbf{AG}  & \textbf{DSTC}   & \textbf{SST-2}     & \textbf{RT} \\ \midrule
\textbf{Before attack} &\textit{0.95}&\textit{0.89}&\textit{0.94}&\textit{0.89}\\
\hline
DeepWordBug&0.93&0.70&0.77&0.79 &0.02&0.21&0.18&0.10 & 0.90 &0.92 & 0.88 & 0.81 \\
HotFlip&0.94&0.80&0.92&0.87 &0.01&0.09&0.02&0.03 & \textbf{0.43} & 0.59 & \textbf{0.44} & \textbf{0.46} \\
PWWS&0.93&0.74&0.78&0.79 &0.02&0.18&0.17&0.11 & 0.79 & 0.86 & \underline{0.64} & 0.60 \\
TextBugger&0.93&0.78&0.89&0.83 &0.02&0.12&0.06&0.06 & 0.85 & 0.84 &0.68 & 0.63 \\
\textit{SamplingFool}&0.89&0.64&0.72&0.82 &0.06&0.27&0.23&0.07 & 0.75 & 0.44 & 0.65 & 0.60 \\
\textit{DILMA}&\textbf{0.82}&\textbf{0.57}&\underline{0.57}&\underline{0.77} & \textbf{0.13}&\textbf{0.34}&\underline{0.37}&\underline{0.12} & 0.81 & 0.43 & 0.65 & 0.55 \\
\textit{DILMA with DL}&\underline{0.86}&\underline{0.58}&\textbf{0.53}&\textbf{0.76} &\underline{0.09}&\underline{0.33}&\textbf{0.41}&\textbf{0.13} & \underline{0.73} & \textbf{0.43} & 0.65 & \underline{0.52} \\ 
\midrule
\end{tabular}
\caption{Additional performance scores before (first row) and after different attacks. Our methods are in \textit{italic}. The best values are in \textbf{bold}, the second best values are \underline{underscored}.}
\label{table:accuracy_table}
\end{table*}

\begin{table*}[]
\centering
\begin{tabular}{lllllllll}
\toprule
\multicolumn{1}{l}{} & 
\multicolumn{2}{c}{\textbf{AG}}&  \multicolumn{2}{c}{\textbf{DSTC}}& 
\multicolumn{2}{c}{\textbf{SST-2}}&  \multicolumn{2}{c}{\textbf{RT}} \\
\cmidrule(r){2-3} \cmidrule(r){4-5} \cmidrule(r){6-7} \cmidrule(r){8-9}
\multicolumn{1}{l}{} &Dist&Ent&Dist&Ent&Dist&Ent&Dist&Ent\\
\hline
\multicolumn{1}{l}{\textbf{Before the attack:}} &0.69&9.81&0.34&7.48&0.72&8.64&0.71&9.30\\ \hline
DeepWordBug&0.75 $\uparrow$&9.95 $\uparrow$&0.46 $\uparrow$&7.90 $\uparrow$&0.76 $\uparrow$&8.72 $\uparrow$&0.75$\uparrow$&9.37 $\uparrow$\\
HotFlip&0.69&9.82 $\uparrow$&0.38 $\uparrow$&7.61 $\uparrow$&0.72&8.64&0.71&9.30\\
PWWS&0.73 $\uparrow$&9.91 $\uparrow$&0.40 $\uparrow$&7.75 $\uparrow$&0.74 $\uparrow$&8.68 $\uparrow$&0.73 $\uparrow$&9.34 $\uparrow$\\
TextBugger&0.74 $\uparrow$&9.99 $\uparrow$&0.42&7.80 $\uparrow$&0.75 $\uparrow$&8.71 $\uparrow$&0.73 $\uparrow$&9.36 $\uparrow$\\
\textit{SamplingFool (ours)} &0.71 $\uparrow$&9.98 $\uparrow$&0.43 $\uparrow$&7.76 $\uparrow$&0.76&8.84  $\uparrow$&0.74 $\uparrow$&9.38 $\uparrow$\\
\textit{DILMA (ours)}&0.63 $\downarrow$&9.53 $\downarrow$&0.41 $\uparrow$&7.73 $\uparrow$&0.73 $\uparrow$&8.77$\uparrow$&0.72 $\uparrow$&9.34 $\uparrow$\\
\textit{DILMA with DL (ours)}&0.69&9.89 $\uparrow$&0.41 $\uparrow$&7.72 $\uparrow$&0.74 $\uparrow$&8.78$\uparrow$&0.73 $\uparrow$&9.36 $\uparrow$\\
\hline
\end{tabular}
\caption{Dist-2 and Ent-2 results after an attack. $\uparrow$ / $\downarrow$ shows, if a metric was increased or decreased. Results for the initial sequences are in the first row. }
\label{table:dist_ent_results}
\end{table*}

\subsection{Linguistic evaluation} 
To measure the linguistic features of the generated adversarial sentences we calculated several metrics, which can be divided into two groups: showing how far the generated samples are from the original ones and demonstrating the linguistic quality of adversarial examples.

% {\bf Semantic similarity} was automatically evaluated by using F1-BERTScore~\cite{bert-score}. BERTScore is a composite metric that leverages the pre-trained contextual embeddings from BERT and aligns words in candidate and reference sentences by cosine similarity. It has been shown to correlate with human judgment in sentence-level and system-level evaluation. Table~\ref{table:linguistic_results} provides mean F1 BERTScore estimates. 
% Our attacks provide more interesting and diverse values w.r.t. BERTscore; thus, they are better in terms of diversity of the generated sequences.

% \begin{table}[]
% \centering
% \begin{tabular}{lcccc}
% \toprule
% \multicolumn{1}{l}{} & \multicolumn{1}{c}{\textbf{AG}}   &  \multicolumn{1}{c}{\textbf{DSTC}}   &  \multicolumn{1}{c}{\textbf{SST-2}}     &  \multicolumn{1}{c}{\textbf{RT}} \\ \midrule
% DeepWordBug&0.61&0.74&0.51&0.68\\
% HotFlip&\textbf{0.97}&\textbf{0.90}&\textbf{0.98}&\textbf{0.98}\\
% PWWS&\underline{0.72}&\underline{0.82}&0.72&\underline{0.83}\\
% TextBugger&0.69&0.81&\underline{0.76}&0.80\\
% \textit{SamplingFool}&0.65&0.73&0.52&0.67\\
% \textit{DILMA}&0.49&0.70&0.47&0.64\\
% \textit{DILMA with DL}&0.71&0.72&0.48&0.68\\
% \hline
% \end{tabular}
% \caption{Semantic similarity between original and generated sentences. Our methods are in \textit{italic}. The values related to closest texts are in \textbf{bold}, the second closest values are \underline{underscored}.}
% \label{table:linguistic_results}
% \end{table}

These results show that the generated sentences are not very different from the original ones. The same conclusion also follows from the morphological and syntactic analyses. 

We evaluate a {\bf similarity of part-of-speech (POS) annotations} between original and adversarial sentences using the Jaccard index (the intersection of two sets over the union of two sets) between sets containing the POS tags of words from the original $A$ and adversarial $B$ sentences.  Duplicated POS-tags in the sets are allowed. Same way, the {\bf similarity of syntax annotations} is estimated as the Jaccard index of dependency relation tags. We use the Stanza toolkit \cite{qi2020stanza} for POS-tagging and dependency parsing.  The detailed report about morphological and syntactic similarities can be found in the Appendix.

{\bf Diversity} across the generated sentences was evaluated by two measures. The first measure, \emph{Dist-k}, is the total number of distinct $k$-grams divided by the total number of produced tokens in all the generated sentences~\cite{li2016diversity}.  
The second measure, \emph{Ent-k}~\cite{zhang2018generating}  considers that infrequent $k$-grams contribute more to diversity than frequent ones.  Table~\ref{table:dist_ent_results} presents the results for $k=2$. We choose this value as being more indicative to estimate the diversity. A table for other values of $k$ can be found in the Appendix.
As we can see, presented methods preserve the lexical diversity.

% \begin{table}[]
% \centering
% \begin{tabular}{lccc}
% \toprule
% Attack & Accuracy & Similar & Artificial\\
% \midrule
% DeepWordBug & 0.71  &  4.52  & 2.55 \\ 
% TextBugger & 0.70  & 4.53  & 2.4 \\ 
% HotFlip & 0.72  & 4.71  & 2.35\\
% PWWS & 0.70 & 3.77  & 2.33 \\
% \textit{DILMA (ours)} & 0.70  & 3.25 & 2.01 \\
% \midrule
% Original & 0.80  & 5.00  & 2.24 \\

% \bottomrule 
% \end{tabular}
% \caption{Human evaluation results for SST2}
% \label{table:he_res}
% \end{table}

\subsection{Human evaluation} % 
\label{sec:evaluation}

We conducted a human evaluation to understand how comprehensive DILMA adversarial perturbations are and to compare DILMA to other approaches. We used a sample from the SST-2 dataset of size $200$, perturbed with five different attacks.

We recruited crowd workers from a crowdsourcing platform to estimate the accuracies of the methods we compare. Given a sentence and the list of classes, the workers were asked to define a class label. We used original sentences to control workers' performance and estimated the results on adversarial sentences. Each sentence was shown to 5 workers. We used the majority vote as the final label. For all different attacks we achieve similar performances at about 0.70 $\pm$ 0.02, with HotFlip scoring 0.72 and DILMA scoring 0.7.
 
% {\bf Accuracy}. Choose a label for the perturbed sentence.
% {\bf Is similar to source?} Give a score of how similar a given perturbed sentence is to a source one.
% {\bf Is artificial?} Give a score of how likely it is that some of the words in the sentence are corrupted or modified by a computer program.

% For each task, each sentence was shown to 5 workers. For the first task, we used the majority vote as the final label. For the last two tasks, we used a scale from 1 to 5: the higher the score, the more similar two sentences are, or the more likely the sentence is artificial. For these tasks, we averaged the scores.

% Table~\ref{table:he_res} shows metrics for original and modified sequences. In comparison to other methods, sentences perturbed with DILMA are more diverse, less similar to source ones, and sound more natural. 
% The accuracy values are on par with other methods, meaning that the overall sentiment is preserved in perturbed sentences.

\section{Conclusion}

Constructing adversarial attacks for natural language processing is a challenging problem due to the discrete nature of input data and non-differentiability of the loss function. Our idea is to combine sampling from a masked language model (MLM) with tuning of its parameters to produce truly adversarial examples. To tune parameters of the MLM we use a loss function based on two differentiable surrogates -- for a distance between sequences and for an attacked classifier. This results in the proposed DILMA approach. If we only sample from the MLM, we obtain a simple baseline SamplingFool.

In order to estimate the efficiency of adversarial attacks on categorical sequences we have proposed a metric combining WER and the accuracy of the target classifier. For considered diverse NLP datasets, our approaches demonstrate a good performance. 
Moreover, in contrast to competing methods, our approaches win over common strategies used to defend from adversarial attacks.
Human and linguistic evaluation also show the adequacy of the proposed attacks.

% \section{Broader impact}

% One of the biggest problems with Deep Learning (DL) models is that they lack robustness. An excellent example is provided by adversarial attacks: a small perturbation of a target image fools a DL classifier, which predicts a wrong label for the perturbed image. The original gradient-based approach to adversarial attacks is not suitable for categorical sequences (e.g. NLP data), as the gradients are not easy to obtain. Existing approaches to adversarial attacks for categorical sequences often fail to keep the meaning and semantics of the original sequence. 

% Our approach is better at this task, as it leverages modern language models like BERT. With the help of our approach, one can generate meaningful adversarial sequences that are persistent against adversarial training and defense strategies based on detection of adversarial examples. This approach poses an important question to society: can we delegate the processing of sequential data to AI? An example of a malicious use of such an approach would be an attack on a model that detects Fake news in major social networks, as an adversarial change undetectable to a human eye can render that model useless.

% We hope that our work will have a broad impact, as we have made our code and all details of our experiments available to ML community. Moreover, it is easy to use, because it requires only a masked language model to work.

\bibliography{anthology,acl2021}
\bibliographystyle{acl_natbib}

\end{document}